\documentclass[10pt,twocolumn,letterpaper]{article}

\usepackage{cvpr}
\usepackage{times}
\usepackage{epsfig}
\usepackage{graphicx}
\usepackage{amsmath}
\usepackage{amssymb}
\usepackage{tabulary}
\usepackage{bbm}

\usepackage[pagebackref=true,breaklinks=true,letterpaper=true,colorlinks,bookmarks=false]{hyperref}

\cvprfinalcopy 

\def\cvprPaperID{1267} 

\ifcvprfinal\fi
\begin{document}

\title{DenseCap: Fully Convolutional Localization Networks for Dense Captioning}

\newcommand*\samethanks[1][\value{footnote}]{\footnotemark[#1]}

\author{
  Justin Johnson\thanks{Indicates equal contribution.} \hspace{3.1pc}
  Andrej Karpathy\samethanks \hspace{3.1pc}
  Li Fei-Fei \\
  Department of Computer Science, Stanford University \\
  {\tt\small \{jcjohns, karpathy, feifeili\}@cs.stanford.edu}
}


\maketitle

\begin{abstract}
\noindent We introduce the dense captioning task, which requires a computer vision system to both
localize and describe salient regions in images in natural language. The dense captioning task generalizes object detection 
when the descriptions consist of a single word, and Image Captioning when one predicted region covers the full image. 
To address the localization and description task jointly we propose a Fully Convolutional Localization Network (FCLN)
architecture that processes an image with a single, efficient forward pass, requires no external regions proposals,
and can be trained end-to-end with a single round of optimization. The architecture is composed of
a Convolutional Network, a novel dense localization layer, and Recurrent Neural Network language model 
that generates the label sequences. We evaluate our network on the Visual Genome dataset, which comprises 
94,000 images and 4,100,000 region-grounded captions. We observe both speed and accuracy improvements over 
baselines based on current state of the art approaches in both generation and retrieval settings.
\end{abstract}

\vspace{-0.15in}
\section{Introduction}
\vspace{-0.05in}

\noindent Our ability to effortlessly point out and describe all aspects of an image relies on a strong semantic
understanding of a visual scene and all of its elements. However, despite numerous 
potential applications, this ability remains a challenge for our state of the art visual recognition systems.
In the last few years there has been significant progress in image classification~\cite{ilsvrc,krizhevsky2012imagenet,zeiler2013visualizing,szegedy2014going}, 
where the task is to assign one label to an image. Further work has pushed these advances along two orthogonal
directions: First, rapid progress in object detection~\cite{overfeat,rcnn,szegedy2014scalable} has identified models that efficiently
identify and label multiple salient regions of an image. Second, recent advances in image captioning \cite{mscococap,mao2014explain,karpathy2014deep,vinyals2014show,xu2015show,donahue2014long,chen2014learning} 
have expanded the complexity of the label space from a fixed set of categories to sequence of
words able to express significantly richer concepts.

\begin{figure}[t]
  \centering
  \includegraphics[width=\linewidth]{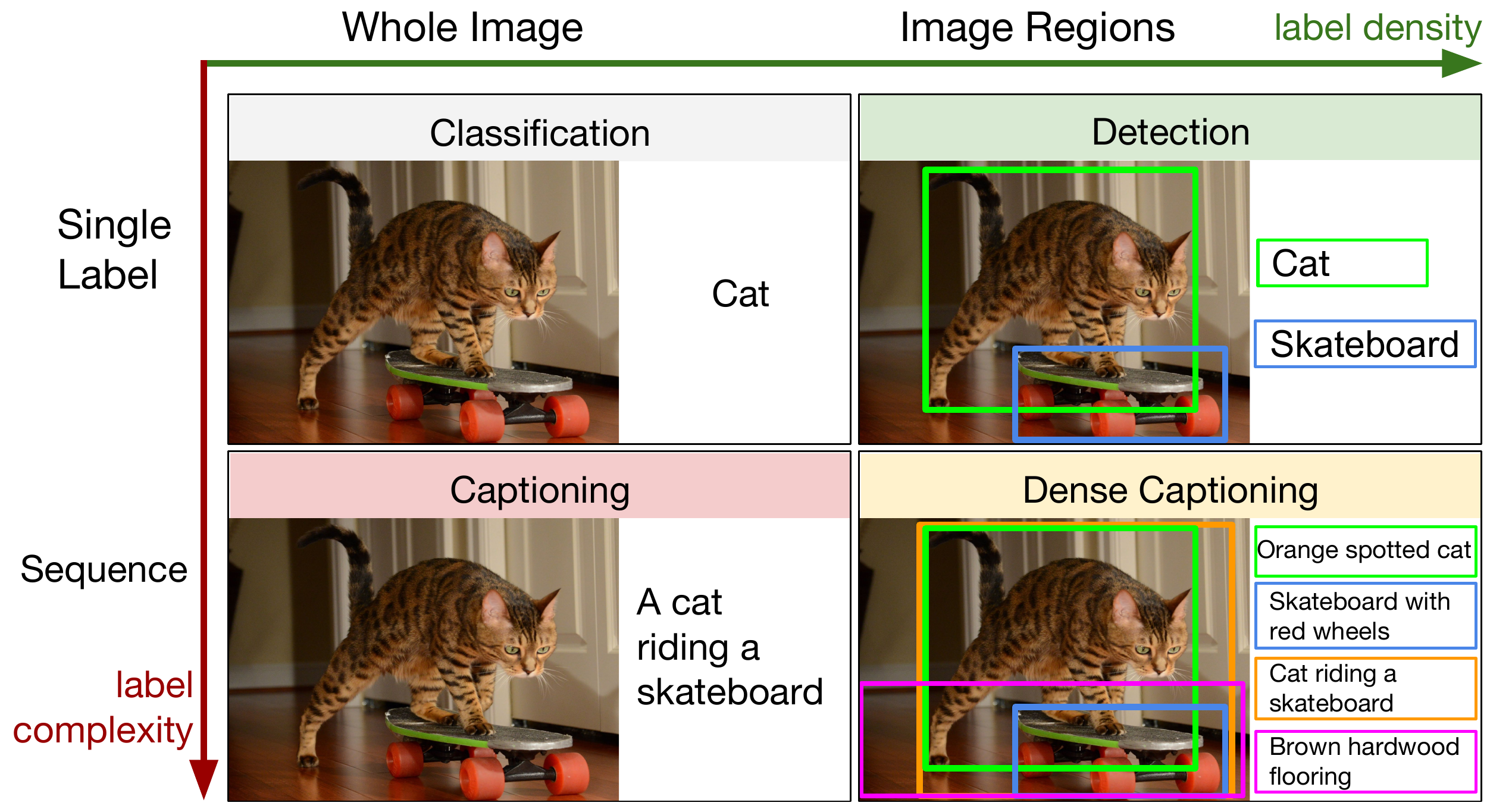}
  \caption{We address the Dense Captioning task (bottom right) by generating
    dense, rich annotations with a single forward pass.}
  \label{fig:pull}
  \vspace{-0.15in}
\end{figure}

However, despite encouraging progress along the label density and label complexity axes, these two 
directions have remained separate. In this work we take a step towards unifying these two inter-connected
tasks into one joint framework. First, we introduce the dense captioning task (see Figure \ref{fig:pull}),
which requires a model to predict a set of descriptions across regions of an image.
Object detection is hence recovered as a special case when the target labels consist of one word, 
and image captioning is recovered when all images consist of one region that spans the full image.

Additionally, we develop a Fully Convolutional Localization Network architecture (FCLN) to address the dense captioning
task. Our model is inspired by recent work in image captioning \cite{vinyals2014show,karpathy2014deep,mao2014explain,donahue2014long,chen2014learning}
in that it is composed of a Convolutional Neural Network followed by a Recurrent Neural Network language
model. However, drawing on work in object detection \cite{fasterrcnn}, our second core contribution is to introduce a
new dense localization layer. This layer is fully differentiable and can be inserted into any neural network that 
processes images to enable region-level training and predictions. Internally, the localization layer predicts 
a set of regions of interest in the image and then uses bilinear interpolation \cite{stn,draw} to smoothly extract the activations 
inside each region.

We evaluate the model on the large-scale Visual Genome dataset, which contains 94,000 images 
and 4,100,000 region captions.  Our results show both performance and speed improvements over approaches based 
on previous state of the art. We make our code and data publicly available to support further progress on the 
dense captioning task.

\section{Related Work}
\vspace{-0.05in}

\noindent Our work draws on recent work in object detection, image captioning, and soft spatial attention
that allows downstream processing of particular regions in the image.

\noindent \textbf{Object Detection.} Our core visual processing module is a Convolutional Neural Network (CNN)
\cite{lecun1998gradient,krizhevsky2012imagenet}, which has emerged as a powerful model for
visual recognition tasks \cite{ilsvrc}. The first application of these models to dense prediction tasks was
introduced in R-CNN \cite{rcnn}, where each region of interest was processed independently. Further work has focused
on processing all regions with only single forward pass of the CNN \cite{spp,fastrcnn}, and on eliminating explicit
region proposal methods by directly predicting the bounding boxes either in the image coordinate system
\cite{szegedy2014scalable,erhan2014scalable}, or in a fully convolutional~\cite{long2014fully} and hence position-invariant
settings \cite{overfeat,fasterrcnn,redmon2015you}. Most related to our approach is the work of Ren \etal
\cite{fasterrcnn} who develop a region proposal network (RPN) that regresses from anchors to regions of interest.
However, they adopt a 4-step optimization process, while our approach does not require training pipelines.
Additionally, we replace their RoI pooling mechanism with a differentiable, spatial soft attention 
mechanism \cite{stn,draw}. In particular, this change allows us to backpropagate through the region
proposal network and train the whole model jointly.

\noindent \textbf{Image Captioning.} Several pioneering approaches have explored the task of describing 
images with natural language
\cite{barnard2003matching,kulkarni2011baby,farhadi2010every,ordonez,socher2010connecting,sochergrounded,kuznetsova2013generalizing,jia2011learning}.
More recent approaches based on neural networks have adopted Recurrent Neural Networks (RNNs)~\cite{rnn,lstm} as the
core architectural element for generating captions. These models have previously been
used in language modeling~\cite{bengio2003neural,graves2013generating,mikolov2010recurrent,sutskever2011generating},
where they are known to learn powerful long-term interactions~\cite{rnnvis}. Several recent approaches
to Image Captioning~\cite{mao2014explain,karpathy2014deep,vinyals2014show,donahue2014long,chen2014learning,kiros2014unifying,fang2014captions}
rely on a combination of RNN language model conditioned on image information. A recent related approach is the work
of Xu \etal \cite{xu2015show} who use a soft attention mechanism \cite{softattend} over regions of the input image 
with every generated word. Our approach to spatial attention is more general in that the network can 
process arbitrary affine regions in the image instead of only discrete grid positions in an intermediate conv volume.
However, for simplicity, during generation we follow Vinyals \etal \cite{vinyals2014show}, where the visual 
information is only passed to the language model once on the first time step.

Finally, the metrics we develop for the dense captioning task are inspired by metrics developed for image 
captioning \cite{cider,meteor,mscococap}.

\begin{figure*}[ht!]
  \centering
  \includegraphics[width=0.9\textwidth]{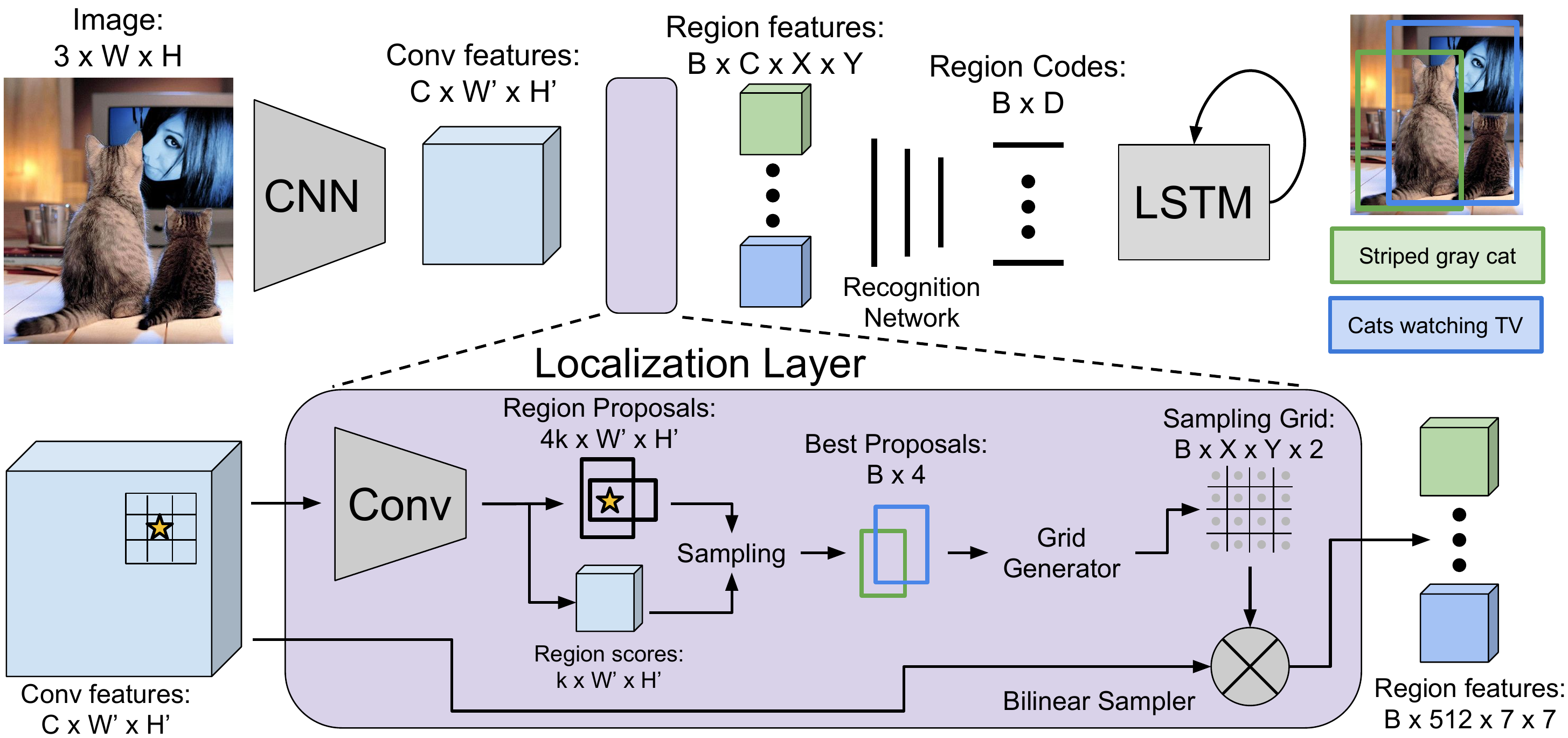}
   \caption{Model overview. An input image is first processed a CNN. The Localization Layer
   proposes regions and smoothly extracts a batch of corresponding activations using bilinear interpolation.
   These regions are processed with a fully-connected recognition network and described with an RNN language model.
   The model is trained end-to-end with gradient descent.}
\label{fig:net}
\vspace{-0.1in}
\end{figure*}

\section{Model}
\vspace{-0.05in}


\noindent \textbf{Overview.}
Our goal is to design an architecture that jointly localizes regions of interest and then describes each
with natural language. The primary challenge is to develop a model that supports end-to-end training with
a single step of optimization, and both efficient and effective inference. Our
proposed architecture (see Figure~\ref{fig:net}) draws on architectural elements present in recent work on 
object detection, image captioning and soft spatial attention to simultaneously address these design constraints.



\noindent In Section \ref{sec:model} we first describe the components of our model. Then in Sections \ref{sec:loss}
and \ref{sec:protocol} we address the loss function and the details of training and inference.

\subsection{Model Architecture}
\label{sec:model}

\vspace{-0.05in}
\subsubsection{Convolutional Network}
\vspace{-0.05in}

We use the VGG-16 architecture \cite{simonyan2014very} for its state-of-the-art performance \cite{ilsvrc}.
It consists of 13 layers of $3\times3$ convolutions interspersed with 5 layers of $2\times2$
max pooling. We remove the final pooling layer, so an input image of shape $3 \times W \times H$
gives rise to a tensor of features of shape $C \times W' \times H'$ where $C=512$,
$W'=\left\lfloor\frac{W}{16}\right\rfloor$, and $H'=\left\lfloor\frac{H}{16}\right\rfloor$.
The output of this network encodes the appearance of the image
at a set of uniformly sampled image locations, and forms the input to the localization layer.

\vspace{-0.05in}
\subsubsection{Fully Convolutional Localization Layer}
\vspace{-0.05in}

The localization layer receives an input tensor of activations, identifies spatial regions of interest
and smoothly extracts a fixed-sized representation from each region. Our approach is based on that of
Faster R-CNN \cite{fasterrcnn}, but we replace their RoI pooling mechanism \cite{fastrcnn}
with bilinear interpolation \cite{stn}, allowing our model to propagate gradients backward through
the coordinates of predicted regions. This modification opens up the possibility of
predicting affine or morphed region proposals instead of bounding boxes \cite{stn},
but we leave these extensions to future work.

\noindent \textbf{Inputs/outputs}. The localization layer accepts a tensor of activations of size
$C \times W' \times H'$. It then internally selects $B$ regions of interest and returns three output
tensors giving information about these regions:

\begin{enumerate}
  \item \textbf{Region Coordinates}: A matrix of shape $B\times 4$ giving bounding box
    coordinates for each output region.
  \item \textbf{Region Scores}: A vector of length $B$ giving a confidence score for each output
    region. Regions with high confidence scores are more likely to correspond to ground-truth
    regions of interest.
  \item \textbf{Region Features}: A tensor of shape $B\times C\times X\times Y$ giving features
    for output regions; is represented by an $X\times Y$ grid of $C$-dimensional features.
\end{enumerate}

\noindent \textbf{Convolutional Anchors}. Similar to Faster R-CNN \cite{fasterrcnn}, our localization layer
predicts region proposals by regressing offsets from a set of translation-invariant anchors. In particular,
we project each point in the $W'\times H'$ grid of input features back into the $W\times H$ image plane, and consider
$k$ anchor boxes of different aspect ratios centered at this projected point. For each of these $k$ anchor boxes, 
the localization layer predicts a confidence score and four scalars regressing from the anchor to the predicted
box coordinates. These are computed by passing the input feature map through a $3\times 3$ convolution
with 256 filters, a rectified linear nonlinearity, and a $1\times 1$ convolution with $5k$ filters.
This results in a tensor of shape $5k \times W' \times H'$ containing scores and offsets for all anchors.

\noindent \textbf{Box Regression}.
\label{sec:box-regression}
We adopt the parameterization of \cite{fastrcnn} to regress from anchors to the region proposals.
Given an anchor box with center $(x_a, y_a)$, width $w_a$, and height $h_a$, our model predicts
scalars $(t_x, t_y, t_w, t_h)$ giving normalized offsets and log-space scaling transforms, so that
the output region has center $(x, y)$ and shape $(w, h)$ given by

\vspace{-0.2in}
\begin{align}
  x &= x_a + t_x w_a &
  y &= y_a + t_y h_a \\
  w &= w_a \exp(t_w) &
  h &= h_a \exp(h_w)
\end{align}
\vspace{-0.2in}

\noindent \textbf{Box Sampling}.
Processing a typical image of size $W=720,H=540$ with $k=12$ anchor boxes gives rise to 17,280
region proposals. Since running the recognition network and the language model for all proposals 
would be prohibitively expensive, it is necessary to subsample them.

At training time, we follow the approach of \cite{fasterrcnn} and sample a minibatch containing
$B=256$ boxes with at most $B/2$ positive regions and the rest negatives. A region is positive
if it has an intersection over union (IoU) of at least $0.7$ with some ground-truth region;
in addition, the predicted region of maximal IoU with each ground-truth region is positive.
A region is negative if it has IoU $<0.3$ with all ground-truth regions.
Our sampled minibatch contains $B_P \leq B/2$ positive regions and $B_N = B - B_P$ negative
regions, sampled uniformly without replacement from the set of all positive and all negative
regions respectively.

At test time we subsample using greedy non-maximum suppression (NMS) based on
the predicted proposal confidences to select the $B=300$ most confident propoals.

The coordinates and confidences of the sampled proposals are collected into
tensors of shape $B\times 4$ and $B$ respectively, and are output from the localization layer.

\noindent \textbf{Bilinear Interpolation.}
After sampling, we are left with region proposals of varying sizes and aspect ratios. In order to
interface with the full-connected recognition network and the RNN language model, we must extract
a fixed-size feature representation for each variably sized region proposal.

To solve this problem, Fast R-CNN \cite{fastrcnn} proposes an RoI pooling layer where each region
proposal is projected onto the $W'\times H'$ grid of convolutional features and divided into a
coarse $X\times Y$ grid aligned to pixel boundaries by rounding. Features are max-pooled
within each grid cell, resulting in an $X\times Y$ grid of output features.

The RoI pooling layer is a function of two inputs: convolutional features and region proposal
coordinates. Gradients can be propagated backward from the output features to the input features,
but not to the input proposal coordinates. To overcome this limitation, we replace the RoI pooling
layer with with bilinear interpolation \cite{draw,stn}.

Concretely, given an input feature map $U$ of shape $C\times W' \times H'$ and a region proposal,
we interpolate the features of $U$ to produce an output feature map $V$ of shape $C\times X\times Y$.
After projecting the region proposal onto $U$ we follow \cite{stn} and compute a \textit{sampling grid} 
$G$ of shape $X\times Y \times 2$ associating each element of $V$
with real-valued coordinates into $U$. If $G_{i,j} = (x_{i,j}, y_{i,j})$ then $V_{c, i, j}$ should be
equal to $U$ at $(c, x_{i,j}, y_{i,j})$; however since $(x_{i,j},y_{i,j})$ are real-valued, we convolve
with a sampling kernel $k$ and set

\vspace{-0.2in}
\begin{equation}
  V_{c, i, j} = \sum_{i'=1}^W\sum_{j'=1}^H U_{c,i',j'}k(i' - x_{i,j})k(j' - y_{i,j}).
\end{equation}
\vspace{-0.15in}

We use bilinear sampling, corresponding to the kernel $k(d) = \max(0, 1-|d|)$.
The sampling grid is a linear function of the proposal coordinates, so gradients can be
propagated backward into predicted region proposal coordinates.
Running bilinear interpolation to extract features for all sampled regions gives a tensor of shape
$B\times C\times X\times Y$, forming the final output from the localization layer.

\vspace{-0.05in}
\subsubsection{Recognition Network}
\vspace{-0.05in}

The recognition network is a fully-connected neural network that processes region features from
the localization layer. The features from each region are flattened into a vector and passed through
two full-connected layers, each using rectified linear units and regularized using Dropout.
For each region this produces a code of dimension $D=4096$ that compactly encodes its visual appearance.
The codes for all positive regions are collected into a matrix of shape $B\times D$ and passed to the RNN
language model.

In addition, we allow the recognition network one more chance to refine the confidence and position of 
each proposal region. It outputs a final scalar confidence of each proposed region and four scalars
encoding a final spatial offset to be applied to the region proposal. These two outputs are computed as
a linear transform from the $D$-dimensional code for each region. The final box regression uses the same
parameterization as Section~\ref{sec:box-regression}.

\vspace{-0.05in}
\subsubsection{RNN Language Model}
\vspace{-0.05in}

Following previous work \cite{mao2014explain,karpathy2014deep,vinyals2014show,donahue2014long,chen2014learning}, 
we use the region codes to condition an RNN language model \cite{graves2013generating,mikolov2010recurrent,sutskever2011generating}.
Concretely, given a training sequence of tokens $s_1, \ldots, s_T$, we feed the RNN $T+2$ word vectors $x_{-1}, x_0, x_1, \ldots, x_T$, 
where $x_{-1} = \text{CNN}(I)$ is the region code encoded with a linear layer and followed by a ReLU non-linearity, 
$x_0$ corresponds to a special START token, and $x_t$ encode each of the tokens $s_t$, $t = 1,\ldots,T$. 
The RNN computes a sequence of hidden states $h_t$ and output vectors $y_t$ using a recurrence 
formula $h_t, y_t = f(h_{t-1}, x_t)$ (we use the LSTM \cite{lstm} recurrence). The vectors $y_t$ have size $|V| + 1$ 
where $V$ is the token vocabulary, and where the additional one is for a special END token. 
The loss function on the vectors $y_t$ is the average cross entropy, where the targets at 
times $t = 0, \ldots, T-1$ are the token indices for $s_{t+1}$, and the target at $t=T$ is the END token. 
The vector $y_{-1}$ is ignored. Our tokens and hidden layers have size 512.

At test time we feed the visual information $x_{-1}$ to the RNN. At each time step we sample the most likely
next token and feed it to the RNN in the next time step, repeating the process until the special END token is sampled.

\vspace{-0.05in}
\subsection{Loss function}
\label{sec:loss}
\vspace{-0.05in}

\noindent During training our ground truth consists of positive boxes and descriptions.
Our model predicts positions and confidences of sampled regions twice: in the localization layer and again
in the recognition network. We use binary logistic lossses for the confidences trained on sampled positive and negative regions.
For box regression, we use a smooth L1 loss in transform coordinate space similar to \cite{fasterrcnn}.
The fifth term in our loss function is a cross-entropy term at every time-step of the language model.

We normalize all loss functions by the batch size and sequence length in the RNN. We searched over an effective
setting of the weights between these contributions and found that a reasonable setting is to use a weight of 0.1
for the first four criterions, and a weight of 1.0 for captioning.

\begin{figure*}[t]
  \centering
  \includegraphics[width=\textwidth]{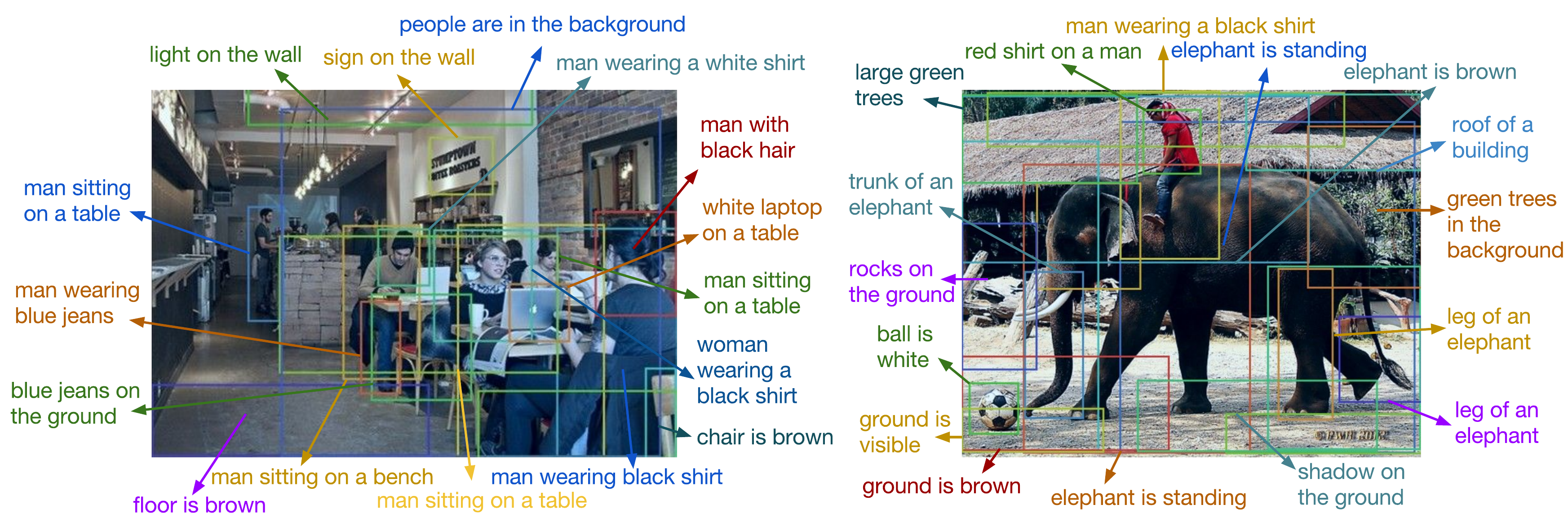} \\*[1mm]
  \includegraphics[width=0.9\textwidth]{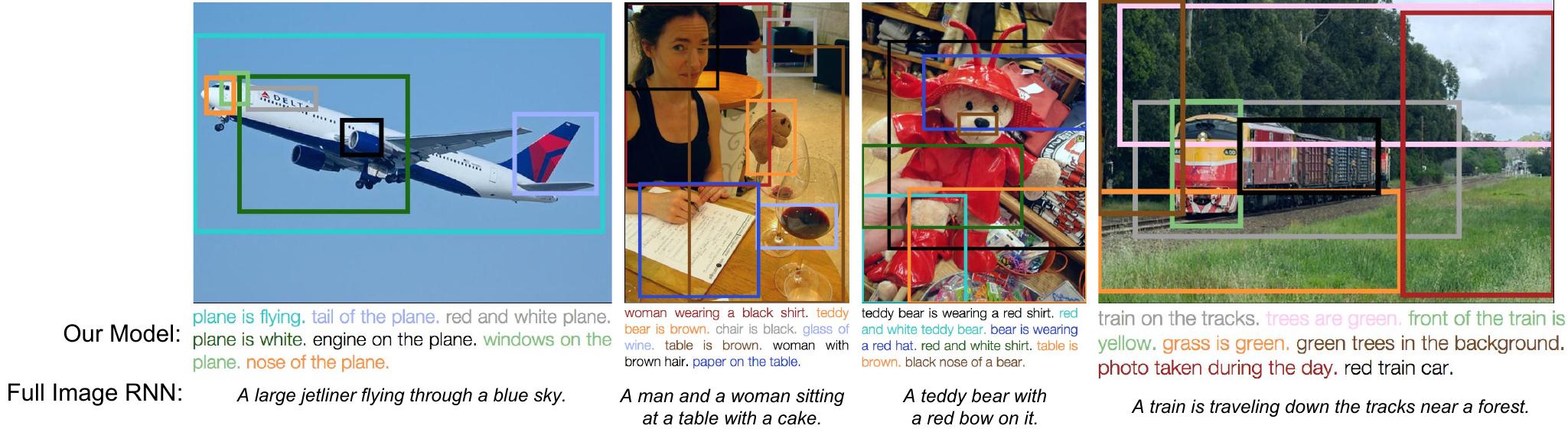}
  \caption{Example captions generated and localized by our model on test images. We render the top few most confident predictions. On the bottom row we additionally contrast the amount of information our model generates compared to the Full image RNN.}
\label{fig:results}
\end{figure*}

\vspace{-0.05in}
\subsection{Training and optimization}
\label{sec:protocol}
\label{sec:training}
\vspace{-0.05in}

\noindent We train the full model end-to-end in a single step of optimization. We initialize the CNN with
weights pretrained on ImageNet \cite{ilsvrc} and all other weights from a gaussian with standard deviation
of 0.01. We use stochastic gradient descent with momentum 0.9 to train the weights of the convolutional network, 
and Adam~\cite{adam} to train the other components of the model. We use a learning rate of $1\times10^{-6}$
and set $\beta_1=0.9,\beta_2=0.99$. We begin fine-tuning the layers of the CNN after 1 epoch, and for
efficiency we do not fine-tune the first four convolutional layers of the network.

Our training batches consist of a single image that has been resized so that the longer side has 720 pixels. 
Our implementation uses Torch7~\cite{torch} and \cite{stnbhwd}. One mini-batch runs in approximately 300ms on a
Titan X GPU and it takes about three days of training for the model to converge.

\vspace{-0.05in}
\section{Experiments}
\vspace{-0.05in}

\noindent \textbf{Dataset}. Existing datasets that relate images and natural language either only
include full image captions \cite{mscococap,flickr30k}, or ground words of image captions in regions
but do not provide individual region captions \cite{flickr30kentities}. We perform our experiments
using the Visual Genome (VG) region captions dataset \footnote{Dataset in submission, obtained via 
personal communication. We commit to releasing the relevant parts upon publication.} This dataset contains 
94,313 images and 4,100,413 snippets of text (43.5 per image), each grounded to a region of an image. 
Images were taken from the intersection of MS COCO
and YFCC100M~\cite{thomee2015new}, and annotations were collected on Amazon Mechanical Turk by
asking workers to draw a bounding box on the image and describe its content in text. Example
captions from the dataset include ``cats play with toys hanging from a perch'', ``newspapers are scattered 
across a table'', ``woman pouring wine into a glass'', ``mane of a zebra'', and ``red light''.

\noindent \textbf{Preprocessing}. We collapse words that appear less than 15 times into a special
\verb|<UNK>| token, giving a vocabulary of 10,497 words. We strip referring phrases such 
as ``there is...'', or ``this seems to be a''. For efficiency we discard all annotations with more 
than 10 words (7\% of annotations). We also discard all images that have fewer than 20 or more 
than 50 annotations to reduce the variation in the number of regions per image.
We are left with 87,398 images; we assign 5,000 each to val/test splits and the rest to train.

For test time evaluation we also preprocess the ground truth regions in the validation/test
images by merging heavily overlapping boxes into single boxes with several reference captions.
For each image we iteratively select the box with the highest number of overlapping
boxes (based on IoU with threshold of 0.7), and merge these together (by taking the mean) into a 
single box with multiple reference captions. We then exclude this group and repeat the process.

\vspace{-0.05in}
\subsection{Dense Captioning}
\vspace{-0.05in}
\label{sec:dense-cap}

\noindent In the dense captioning task the model receives a single image and produces a set of
regions, each annotated with a confidence and a caption.


\begin{table*}
\centering
\small
\def\arraystretch{1.1}
\begin{tabulary}{\linewidth}{LCCC|CCC||CCCC}
\hline
& \multicolumn{3}{c|}{\textbf{Language} (METEOR)} 
& \multicolumn{3}{c||}{\textbf{Dense captioning} (AP)} 
& \multicolumn{4}{c}{\textbf{Test runtime} (ms)} \\
\textbf{Region source} & EB & RPN & GT & EB & RPN & GT & Proposals & CNN+Recog & RNN & Total\\
\hline
\hline
Full image RNN \cite{karpathy2014deep} & 0.173 & 0.197 & 0.209 & 2.42 & 4.27 & \textit{14.11} & 210ms & 2950ms & \textbf{10ms}  & 3170ms \\
Region RNN \cite{karpathy2014deep} & 0.221 & 0.244 & 0.272 & 1.07 & 4.26 & \textit{21.90} & 210ms & 2950ms & \textbf{10ms}  & 3170ms  \\
FCLN on EB \cite{fastrcnn} & \textbf{0.264} & \textbf{0.296} & 0.293 & 4.88 & 3.21 & \textit{26.84} & 210ms & \textbf{140ms} & \textbf{10ms}  &  360ms \\
Our model (FCLN) & \textbf{0.264} & 0.273 & \textbf{0.305} & \textbf{5.24} & \textbf{5.39} & \textbf{\textit{27.03}} & \textbf{90ms} & \textbf{140ms} & \textbf{10ms} & \textbf{240ms}\\
\hline
\end{tabulary}
\vspace{0.05in}
\caption{Dense captioning evaluation on the test set of 5,000 images. The language metric is METEOR (high is good), 
our dense captioning metric is Average Precision (AP, high is good), and the test runtime performance for a $720\times600$ image with  
300 proposals is given in milliseconds on a Titan X  GPU (ms, low is good). 
EB, RPN, and GT correspond to EdgeBoxes \cite{ZitnickECCV14edgeBoxes}, Region Proposal Network \cite{fasterrcnn}, and ground truth boxes respectively, used at test time.
Numbers in GT columns (italic) serve as upper bounds assuming perfect localization.}
\label{tab:results}
\vspace{-0.2in}
\end{table*}

\noindent \textbf{Evaluation metrics}. Intuitively, we would like our model to produce both
well-localized predictions (as in object detection) and accurate descriptions (as in image
captioning). 

Inspired by evaluation metrics in object detection \cite{voc,coco} and image captioning \cite{cider}, 
we propose to measure the mean Average Precision (AP) across a range of thresholds for both 
localization and language accuracy. For localization we use intersection over union (IoU) thresholds
.3, .4, .5, .6, .7. For language we use METEOR score thresholds 0, .05, .1, .15, .2, .25. We adopt
METEOR since this metric was found to be most highly correlated with human judgments in settings 
with a low number of references \cite{cider}. We measure the average precision across all pairwise
settings of these thresholds and report the mean AP.

To isolate the accuracy of language in the predicted captions without localization we also merge
ground truth captions across each test image into a bag of references sentences and evaluate
predicted captions with respect to these references without taking into account their spatial position.

\noindent \textbf{Baseline models}. Following Karpathy and Fei-Fei \cite{karpathy2014deep}, 
we train only the Image Captioning model (excluding the localization layer) on individual, resized regions. 
We refer to this approach as a \textit{Region RNN model}. To investigate the differences between 
captioning trained on full images or regions we also train the same model on full images and captions
from MS COCO (\textit{Full Image RNN model}).

At test time we consider three sources of region proposals. First, to establish an upper bound we evaluate
the model on ground truth boxes (GT). Second, similar to \cite{karpathy2014deep} we use 
an external region proposal method to extract 300 boxes for each test image. We use 
EdgeBoxes \cite{ZitnickECCV14edgeBoxes} (EB) due to their strong performance and speed. 
Finally, EdgeBoxes have been tuned to obtain high recall for objects, but our
regions data contains a wide variety of annotations around groups of objects, stuff, etc. 
Therefore, as a third source of test time regions we follow Faster R-CNN \cite{fasterrcnn}
and train a separate Region Proposal Network (RPN) on the VG regions data. This corresponds 
to training our full model except without the RNN language model.

As the last baseline we reproduce the approach of Fast R-CNN \cite{fastrcnn}, where the region proposals
during training are fixed to EdgeBoxes instead of being predicted by the model (\textit{FCLN on EB}). 
The results of this experiment can be found in Table \ref{tab:results}. We now highlight the main takeaways.

\noindent \textbf{Discrepancy between region and image level statistics}. Focusing on the 
first two rows of Table \ref{tab:results}, the Region RNN model obtains consistently stronger results 
on METEOR alone, supporting the difference in the language statistics present on the level of 
regions and images. Note that these models were trained on nearly the same images, but one on 
full image captions and the other on region captions. However, despite the differences in the
language, the two models reach comparable performance on the final metric.

\noindent \textbf{RPN outperforms external region proposals}. In all cases we obtain 
performance improvements when using the RPN network instead of EB regions. The only exception is the
FCLN model that was only trained on EB boxes. Our hypothesis is that
this reflects people's tendency of annotating regions more general than those containing objects. The
RPN network can learn these distributions from the raw data, while the EdgeBoxes method was designed 
for high recall on objects. In particular, note that this also allows our model (FCLN) to outperform
the FCLN on EB baseline, which is constrained to EdgeBoxes during training (5.24 vs. 4.88 and 5.39 vs. 3.21). This is
despite the fact that their localization-independent language scores are comparable, which suggests that our
model achieves improvements specifically due to better localization. Finally, the noticeable
drop in performance of the FCLN on EB model when evaluating on RPN boxes (5.39 down to 3.21) also suggests 
that the EB boxes have particular visual statistics, and that the RPN boxes are likely out of sample for the
FCLN on EB model.

\noindent \textbf{Our model outperforms individual region description}. Our final model performance is
listed under the RPN column as 5.39 AP. In particular, note that in this one cell of Table \ref{tab:results} 
we report the performance of our full joint model instead of our model evaluated on the boxes from the 
independently trained RPN network. Our performance is quite a bit higher than that of the Region RNN
model, even when the region model is evaluated on the RPN proposals (5.93 vs. 4.26). We attribute this
improvement to the fact that our model can take advantage of visual information from the context outside 
of the test regions.

\noindent \textbf{Qualitative results}. We show example predictions of the dense captioning model in 
Figure \ref{fig:results}. The model generates rich snippet descriptions of regions and accurately grounds
the captions in the images. For instance, note that several parts of the elephant are correctly grounded
and described (``trunk of an elephant'', ``elephant is standing'', and both ``leg of an elephant''). The
same is true for the airplane example, where the tail, engine, nose and windows are correctly localized.
Common failure cases include repeated detections (e.g. the elephant is described as standing twice). 

\noindent \textbf{Runtime evaluation}.
Our model is efficient at test time: a $720\times600$ image is
processed in 240ms. This includes running the CNN, computing $B=300$
region proposals, and sampling from the language model for each region.

Table~\ref{tab:results} (right) compares the test-time runtime performance of our model with baselines that
rely on EdgeBoxes. Regions RNN is slowest since it processes each region with an independent forward
pass of the CNN; with a runtime of 3170ms it is more than $13\times$ slower than our method.

FCLN on EB extracts features for all regions after a single forward pass of the CNN.
Its runtime is dominated by EdgeBoxes, and it is $\approx1.5\times$ slower than our method.

Our method takes 88ms to compute region proposals, of which nearly 80ms is spent running NMS to
subsample regions in the Localization Layer. This time can be drastically reduced by using fewer
proposals: using 100 region proposals reduces our total runtime to 166ms.


\begin{table*}
  \small
  \def\arraystretch{1.1}
  \centering
  \begin{tabulary}{\linewidth}{LCCCC|CCCC}
    \hline
    & \multicolumn{4}{c|}{\textbf{Ranking}} & \multicolumn{4}{c}{\textbf{Localization}} \\
    & R@1 & R@5 & R@10 & Med. rank & IoU@0.1 & IoU@0.3 & IoU@0.5 & Med. IoU \\
    \hline \hline
    Full Image RNN \cite{karpathy2014deep}
        & 0.10 & 0.30 & 0.43 & 13 & - & - & - & - \\
    EB + Full Image RNN \cite{karpathy2014deep} 
        & 0.11 & 0.40 & 0.55 & 9 & 0.348 & 0.156 & 0.053 & 0.020 \\ 
    Region RNN \cite{fastrcnn}        & 0.18 & 0.43 & 0.59 & 7 & 0.460 & 0.273 & 0.108 & 0.077 \\
    Our model (FCLN) & \textbf{0.27} & \textbf{0.53} & \textbf{0.67} & \textbf{5}
        & \textbf{0.560} & \textbf{0.345} & \textbf{0.153} & \textbf{0.137} \\
    \hline
  \end{tabulary}
  \vspace{1mm}
  \caption{Results for image retrieval experiments. We evaluate ranking using
    recall at $k$ (R@$K$, higher is better) and median rank of the target image
    (Med.rank, lower is better). We evaluate localization using ground-truth region
    recall at different IoU thresholds (IoU@$t$, higher is better) and median IoU
    (Med. IoU, higher is better). Our method outperforms baselines at both
    ranking and localization.
  }
  \label{tab:retrieval}
\end{table*}

\begin{figure*}
  \hspace{8mm}
  \textbf{GT image} \hspace{12mm}
  \textbf{Query phrases} \hspace{10pc}
  \textbf{Retrieved Images} \\*[-6mm]
  \begin{center}
    \includegraphics[width=0.95\textwidth]{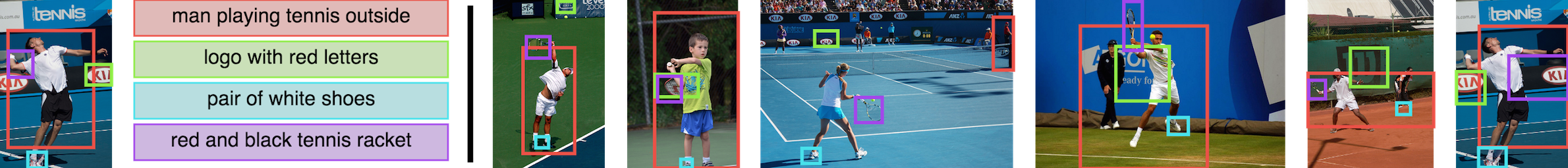} \\*[1mm]
    \includegraphics[width=0.95\textwidth]{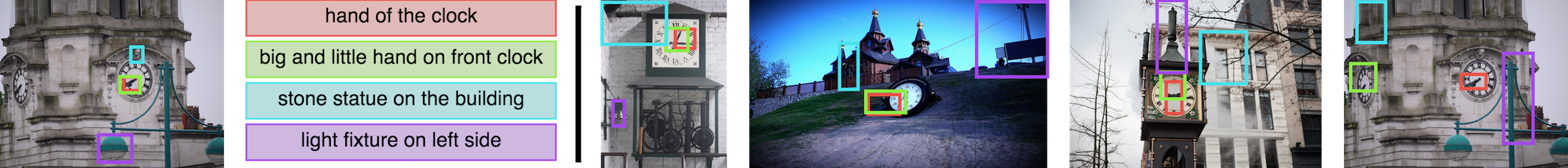} \\*[1mm]
    \includegraphics[width=0.95\textwidth]{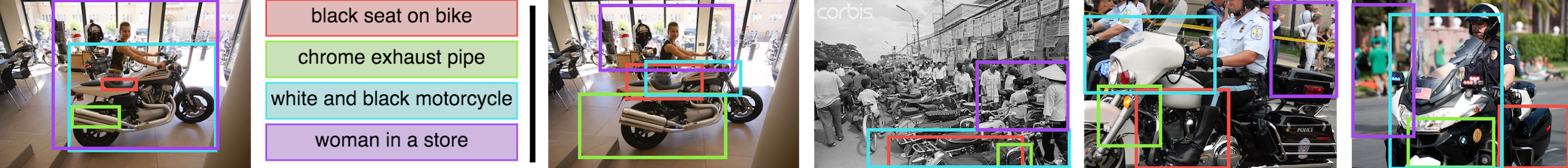} \\*[1mm]
    \includegraphics[width=0.95\textwidth]{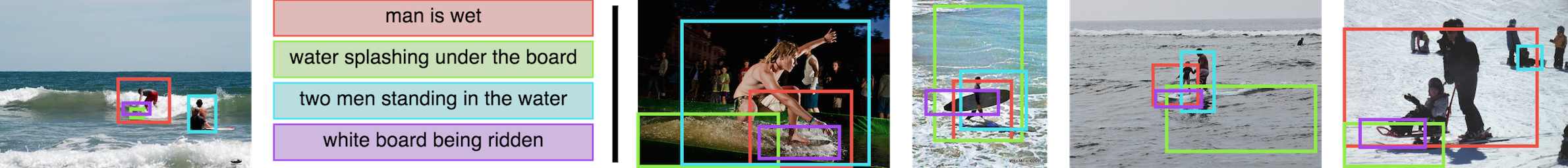} \\*[1mm]
  \end{center}
  \vspace{-2mm}
  \caption{
    Example image retrieval results using our dense captioning model.
    From left to right, each row shows a grund-truth test image, ground-truth region captions describing
    the image, and the top images retrieved by our model using the text of the captions as a query.
    Our model is able to correctly retrieve and localize people, animals, and parts of both
    natural and man-made objects.
  }
  \label{fig:retrieval}
\end{figure*}

\pagebreak
\subsection{Image Retrieval using Regions and Captions}

\noindent In addition to generating novel descriptions, our dense captioning model can support image
retrieval using natural-language queries, and can localize these queries in retrieved images.
We evaluate our model's ability to correctly retrieve images and accurately localize textual queries.

\noindent \textbf{Experiment setup}.
We use 1000 random images from the VG test set for this experiment. We generate
100 test queries by repeatedly sampling four random captions from some image and then 
expect the model to correct retrieve the source image for each query.

\noindent \textbf{Evaluation}.
To evaluate ranking, we report the fraction of queries for which the correct source image
appears in the top $k$ positions for $k\in\{1, 5, 10\}$ (recall at $k$) and the median rank of
the correct image across all queries.

To evaluate localization, for each query caption we examine the image and ground-truth bounding box
from which the caption was sampled. We compute IoU between this ground-truth box and the model's
predicted grounding for the caption. We then report the fraction of query caption for which this overlap
is greater than a threshold $t$ for $t\in\{0.1, 0.3, 0.5\}$ (recall at $t$) and the median IoU
across all query captions.

\begin{figure*}
  \hspace{3mm}
  \begin{minipage}{0.14\textwidth}\itshape\small\centering head of a giraffe \end{minipage}
  \hspace{2mm}
  \begin{minipage}{0.14\textwidth}\itshape\small\centering legs of a zebra \end{minipage}
  \begin{minipage}{0.15\textwidth}\itshape\small\centering red and white sign \end{minipage}
  \hspace{1mm}
  \begin{minipage}{0.15\textwidth}\itshape\small\centering white tennis shoes \end{minipage}
  \begin{minipage}{0.18\textwidth}\itshape\small\centering hands holding a phone \end{minipage}
  \begin{minipage}{0.17\textwidth}\itshape\small\centering front wheel of a bus \end{minipage}
  \\*[-5mm]
  \begin{center}
    \includegraphics[height=0.38\textwidth]{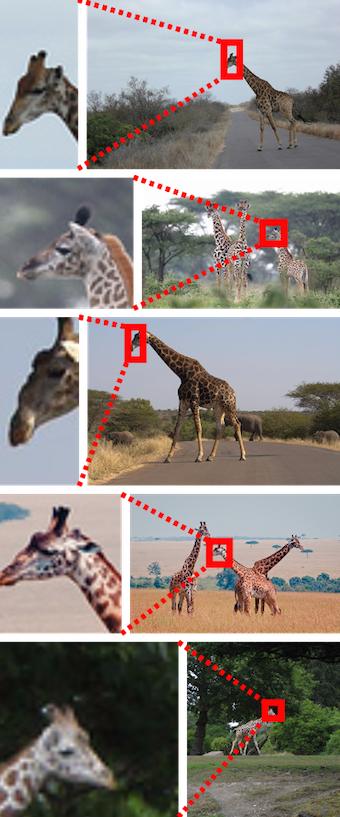} \hspace{0.5mm}
    \includegraphics[height=0.38\textwidth]{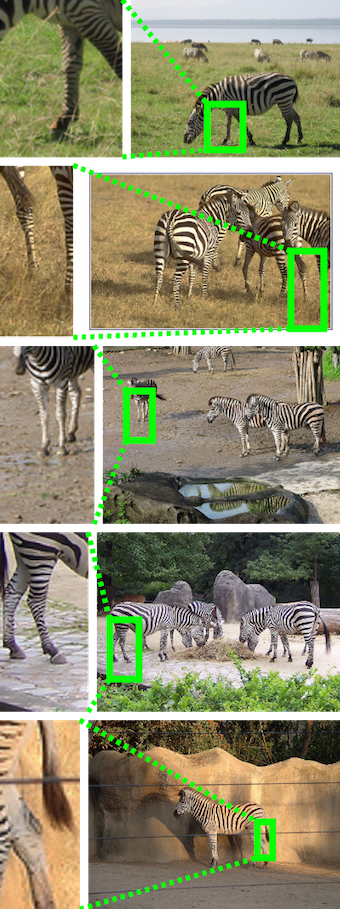} \hspace{0.5mm}
    \includegraphics[height=0.38\textwidth]{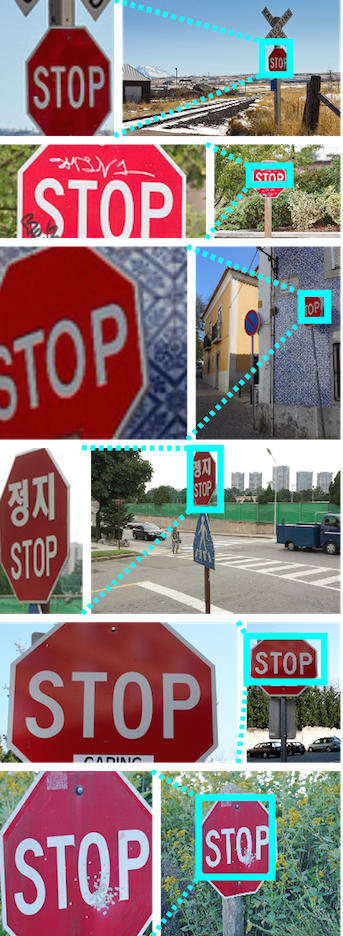} \hspace{0.5mm}
    \includegraphics[height=0.38\textwidth]{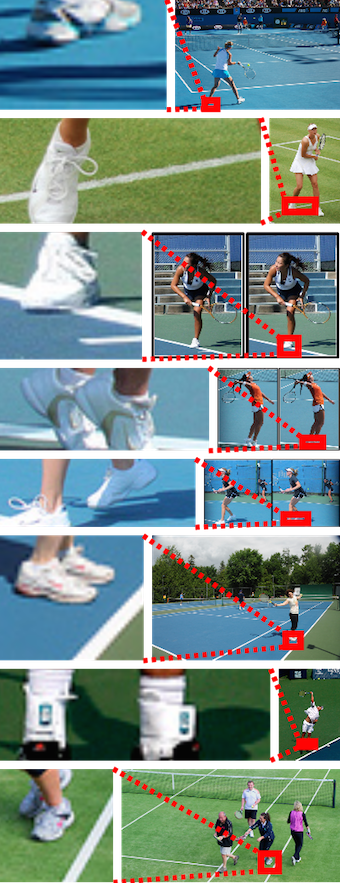} \hspace{0.5mm}
    \includegraphics[height=0.38\textwidth]{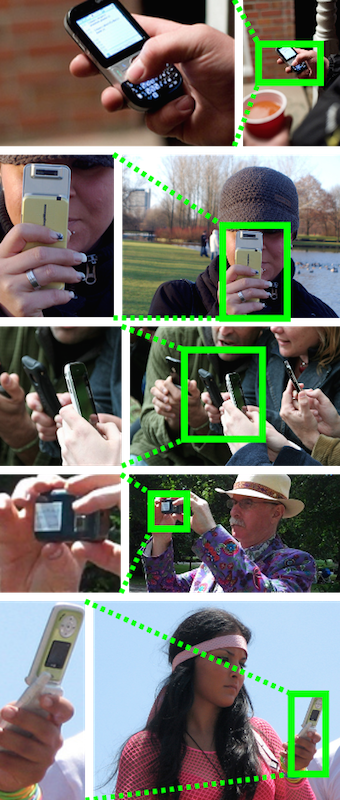} \hspace{0.5mm}
    \includegraphics[height=0.38\textwidth]{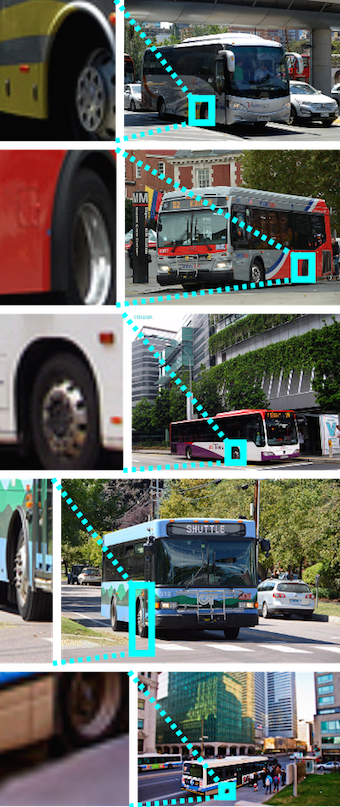}
  \end{center}
  \caption{Example results for open world detection. We use our dense captioning model to localize
    arbitrary pieces of text in images, and display the top detections on the test set for several
    queries.
  }
  \label{fig:open-det}
\end{figure*}

\noindent \textbf{Models}.
We compare the ranking and localization performance of full model with
baseline models from Section~\ref{sec:dense-cap}.

For the Full Image RNN model trained on MS COCO, we compute the probability of generating each
query caption from the entire image and rank test images by mean probability across query captions.
Since this does not localize captions we only evaluate its ranking performance.

The Full Image RNN and Region RNN methods are trained on full MS COCO images and ground-truth
VG regions respectively. In either case, for each query and test image we generate 100 region proposals
using EdgeBoxes and for each query caption and region proposal we compute the probability of generating
the query caption from the region. Query captions are aligned to the proposal of maximal probability,
and images are ranked by the mean probability of aligned caption / region pairs.

The process for the full FCLN model is similar, but uses the top 100 proposals from the localization
layer rather than EdgeBoxes proposals.


\noindent \textbf{Discussion}.
Figure~\ref{fig:retrieval} shows examples of ground-truth images, query phrases describing those
images, and images retrieved from these queries using our model. Our model is
able to localize small objects (``hand of the clock'', ``logo with red letters''),
object parts, (``black seat on bike'', ``chrome exhaust pipe''), people
(``man is wet'') and some actions (``man playing tennis outside'').

Quantitative results comparing our model against the baseline methods is shown in
Table~\ref{tab:retrieval}. The relatively poor performance of the Full Image RNN model (Med. rank 13 vs. 9,7,5) 
may be due to mismatched statistics between its train and test distributions: the model was trained on 
full images, but in this experiment it must match region-level captions
to whole images (Full Image RNN) or process image regions rather than full images
(EB + Full Image RNN).

The Region RNN model does not suffer from  a mismatch between train and test
data, and outperforms the Full Image RNN model on both ranking and localization.
Compared to Full Image RNN, it reduces the median rank from 9
to 7 and improves localization recall at 0.5 IoU from 0.053 to 0.108.

Our model outperforms the Region RNN baseline for both ranking and localization under
all metrics, further reducing the median rank from 7 to 5 and increasing localization
recall at 0.5 IoU from 0.108 to 0.153.

The baseline uses EdgeBoxes which was tuned to localize objects, but not all query phrases
refer to objects. Our model achieves superior results since it learns to propose regions from
the training data.

\noindent \textbf{Open-world Object Detection}
Using the retrieval setup described above, our dense captioning model can also be used to localize
arbitrary pieces of text in images. This enables ``open-world'' object detection,
where instead of committing to a fixed set of object classes at training time we can specify
object classes using natural language at test-time. We show example results for this task in
Figure~\ref{fig:open-det}, where we display the top detections on the test set for several phrases.

Our model can detect animal parts (``head of a giraffe'', ``legs of a zebra'') and also
understands some object attributes (``red and white sign'', ``white tennis shoes'') and interactions
between objects (``hands holding a phone''). The phrase ``front wheel of a bus'' is a failure case:
the model correctly identifies wheels of buses, but cannot distinguish between the front and
back wheel.


\vspace{-0.05in}
\section{Conclusion}
\vspace{-0.05in}

\noindent We introduced the dense captioning task, which requires a model to simultaneously 
localize and describe regions of an image. To address this task we developed the FCLN architecture, 
which supports end-to-end training and efficient test-time performance. Our FCLN architecture is based on
recent CNN-RNN models developed for image captioning but includes a novel, differentiable localization layer
that can be inserted into any neural network to enable spatially-localized predictions. Our experiments
in both generation and retrieval settings demonstrate the power and efficiency of our model 
with respect to baselines related tp previous work, and qualitative experiments show visually pleasing results. In future work
we would like to relax the assumption of rectangular proposal regions and to discard test-time NMS in 
favor of a trainable spatial suppression layer.
\pagebreak

\section{Acknowledgments}
Our work is partially funded by an ONR MURI grant and an Intel research grant.
We thank Vignesh Ramanathan, Yuke Zhu, Ranjay Krishna, and Joseph Lim for helpful comments and
discussion. We gratefully acknowledge the support of NVIDIA Corporation with the donation of the GPUs used for this research.

{\small
\bibliographystyle{ieee}
\bibliography{egbib}
}

\end{document}